# Comparison of techniques for fine-tuning open-weight models for entity extraction from radiology reports

Aawez Mansuri[1], Kush Mehta[2], Mohammadreza Chavoshi[1], Jahanzaib Malik[1], Theodorus Dapamede[1], Frank Li[1], Rohan Isaac[1], Beatrice Brown-Mulry[1], Chiratidzo Rudado Sanyika[1], YoungSeok Jeon[1], Judy W. Gichoya[1], Ali Emami[2], Hari Trivedi[1,*]

[1] Department of Radiology and Imaging Sciences, Emory University School of Medicine, Atlanta, GA, USA

[2] Department of Computer Science, Emory University, Atlanta, GA, USA

* Correspondence: Hari Trivedi, Department of Radiology and Imaging Sciences, Emory University School of Medicine, Atlanta, GA, USA. Email: hari.trivedi@emory.edu

## Abstract

Converting free-text radiology reports into structured labels supports cohort building, quality assurance, and monitoring of clinical imaging models, but the strongest label extractors are hosted proprietary models whose use raises privacy, cost, and reproducibility concerns. We asked whether a fine-tuned open-weight model (Gemma-3-12B) can match GPT-4o at multi-label intracranial hemorrhage (ICH) acuity extraction from non-contrast head-CT reports, and which ingredients matter. Using a 2×2 design, we crossed two adaptation strategies (a discriminative classification head, CH; generative instruction fine-tuning, IFT) with two training-data sources (distillation of real GPT-4o-labeled reports; synthetic reports generated by GPT-4o from real exemplars), across five training sizes, benchmarked on 100 expert-adjudicated reports against GPT-4o and the un-tuned open-weight base. The distilled instruction-tuned model (DIFT) matched GPT-4o (macro-F1 0.845 vs 0.850; $p = 1.000$) and exceeded the base model by 0.178. The decisive factor was the training-data source, not the fine-tuning method: both synthetic-data models failed to exceed the un-tuned open-weight base at any training size and underperformed the distilled models across all acuity classes. Fine-tuning and inference fit within the memory envelope of a single 24 GB consumer GPU. For narrow, high-value clinical label-extraction tasks, distilling real reports, rather than generating synthetic ones, is what closes the gap to a hosted model, enabling a private, low-cost, version-stable on-premises alternative.

## Introduction

Radiology reports are generated as free text, yet much of their secondary value depends on converting that text into discrete, machine-readable labels for cohort assembly, registry curation, quality assurance, and the training and auditing of imaging models. Rule-based keyword systems and, later, encoder-based models such as BERT advanced this task but remain constrained: they are tied to fixed label schemas, require task-specific labeled corpora, and handle long or atypical reports poorly [1-8]. Generative large language models (LLMs) have become an attractive alternative because they relax these constraints, offering long context, flexible instruction-following, and adaptation to new schemas with little task-specific training [9,10].

Within this paradigm, hosted proprietary models such as GPT-4o have generally set the performance ceiling on clinical extraction, and recent frontier general-purpose models now outperform even specialized clinical AI tools on medical benchmarks [11]. On clinical label extraction specifically, open-weight models have tended to trail leading proprietary ones by a small margin [12-14]. Yet routine reliance on a hosted model carries costs that a benchmark table does not capture: protected health information generally cannot leave the institutional perimeter, per-query API charges accumulate at archive scale, and the hosted model is not a fixed instrument. Its behavior can change between releases and versions are deprecated within months, turning validation into a recurring obligation [15-18]. An on-premises, open-weight model adapted to the task keeps data inside the institution, incurs no marginal API fees, and freezes an auditable, version-controlled model.

Two teacher-derived data strategies are commonly used to close the accuracy gap. In distillation, a strong teacher labels real reports and a smaller open-weight student is fine-tuned on those labels [15,19,20]. In a synthetic approach, the teacher instead generates the reports, here conditioned on a few real reports as in-context exemplars so the output mimics authentic reporting, which reduces reliance on real-report corpora and yields a shareable, non-identifiable training set [16,21]. These strategies are not equivalent: broadly imitating a proprietary model is a “false promise,” whereas distilling a single, well-specified behavior can approach teacher-level accuracy on that narrow task [22]. Which strategy succeeds, and how much data it needs, is an empirical question. As we show, the choice of data source matters more than the choice of fine-tuning method.

We study this for a concrete, operationally important target: extracting intracranial hemorrhage (ICH) acuity from non-contrast head-CT reports. ICH is a time-critical emergency in which acuity directly shapes management, and acuity-resolved labels are needed to build cohorts, train and evaluate imaging models, and support continuous post-deployment surveillance, where subgroup-level monitoring requires a steady, private supply of reference labels [3,23-27]. Fine-tuning an open-weight model (Gemma-3-12B) under two data sources and two adaptation strategies, we ask whether it can match GPT-4o and identify which ingredients make that possible. Full design details are given in Methods.

## Methods

### Study design

We conducted a retrospective study to evaluate whether a fine-tuned open-weight LLM could match a proprietary LLM at extracting ICH acuity labels from head-CT radiology reports. We compared two parameter-efficient adaptation strategies applied to a shared open-weight base model (Gemma-3-12B): a discriminative classification head (CH) and generative instruction fine-tuning (IFT). Each was trained under two data sources: distillation from real reports labeled by GPT-4o (yielding DCH and DIFT) and synthetic reports generated by GPT-4o (yielding SCH and SIFT). This 2×2 design crosses adaptation strategy with data source. Acuity was treated as multi-label binary classification across four non-exclusive classes: acute/hyperacute/hyperdense, subacute/isodense, chronic/hypodense, and acute-on-chronic (AOC)/mixed, with hyperacute grouped under acute/hyperdense. A report could carry zero, one, or multiple positive labels. The study design is summarised in Figure 1.

| | Distilled<br>Real reports · GPT-4o labels<br>2,000 reports, class-balanced | Synthetic<br>GPT-4o-generated · 3 real exemplars<br>T = 0.4–0.8, 2,000 class-balanced |
|---|---|---|
| CH<br>Classification head<br>*Sigmoid multi-label head*<br>*LoRA on Q, K, V, O* | DCH<br>Distilled × Classification head | SCH<br>Synthetic × Classification head |
| IFT<br>Instruction fine-tuning<br>*JSON Yes/No output*<br>*LoRA on attention + MLP* | DIFT<br>Distilled × Instruction fine-tuning | SIFT<br>Synthetic × Instruction fine-tuning |

*Figure 1. Study design — 2 × 2 factorial. Two data sources (columns: distilled, real head-CT reports labelled by GPT-4o; synthetic, reports generated by GPT-4o from real exemplars) are crossed with two adaptation strategies (rows: classification head, CH; instruction fine-tuning, IFT), yielding four fine-tuned configurations (DCH, SCH, DIFT, SIFT). All four share Gemma-3-12B as the base model (4-bit NF4 quantisation with LoRA rank 8), are trained at $n \in \{200, 500, 1{,}000, 1{,}500, 2{,}000\}$, and are evaluated on the same 100 held-out manually adjudicated head-CT reports against two zero-shot baselines (GPT-4o and Gemma-3-12B-it).*

## Reference dataset and annotation

Two hundred reports were selectively sampled (enriched for ICH-positive cases, given the low prevalence of positive ICH in unselected reports) and manually annotated and adjudicated by radiologists (Table 1). One hundred reports formed the validation set used during fine-tuning and 100 formed the final hold-out test set, which was evaluated identically by every configuration to ensure strict paired comparisons. The test set was independently annotated by three attending radiologists and adjudicated; the validation set was independently annotated by three residents (without overlap) and verified by an attending radiologist. All reports were de-identified before use with the Stanford de-identifier to remove protected health information [31]. ICH presence was recorded as yes, no, or possible, the possible category capturing reports in which the radiologist expressed diagnostic uncertainty or hedging; acuity labels were assigned only to ICH-positive reports, consistent with a hierarchical scheme in which acuity is extracted downstream of a positive ICH determination.

## Training datasets

Two training datasets were constructed, both using GPT-4o (OpenAI; queried via the API without chain-of-thought) as the upstream labeler or generator. The distilled dataset comprised real, de-identified head-CT reports labeled by GPT-4o, balanced at 500 reports per acuity class (2,000 reports total). The synthetic dataset comprised head-CT reports generated by GPT-4o: for each report, three real reports positive for the target class, together with their labels, were randomly sampled from a pool of 300 reports (disjoint from the validation and test sets) and supplied as in-context exemplars, with the generation temperature drawn uniformly between 0.4 and 0.8 per report; generation was performed per class, and the full prompt is given in the Supplement (Appendix A). Both training datasets consisted exclusively of ICH-positive reports, because acuity is extracted only for reports already determined to be ICH-positive; no ICH-negative reports were used for training. The 19 ICH-negative reports in the test set were retained solely to quantify false-positive (hallucination) behavior. From each dataset, subsets of n = 200, 500, 1,000, 1,500, and 2,000 reports were sampled with a fixed seed (42) for each adaptation strategy, yielding 20 fine-tuned models.

## Fine-tuning

We used Gemma-3-12B (Google DeepMind) as the base model, loaded in 4-bit NF4 quantisation via Unsloth. For CH, a discriminative multi-label sigmoid head was attached to the final hidden state (the pooled last-token representation) of the base model, with LoRA adapters (rank 8, $\alpha = 16$) on the query, key, value, and output attention projections; training used AdamW-8bit (learning rate $2\times10^{-4}$, cosine schedule, 10% warm-up, weight decay 0.05, maximum sequence length 1,024) for up to 40 epochs with early stopping (patience 10) on validation macro-F1. For IFT, we used the chat-templated supervised fine-tuning trainer from TRL with LoRA adapters (rank 8, $\alpha = 8$) on all layers (attention plus MLP), training the model to emit a fixed JSON object with one Yes/No value per acuity class (AdamW-8bit, learning rate $2\times10^{-4}$, linear schedule, 10-step warm-up, weight decay $1\times10^{-3}$, maximum sequence length 2,048; 3 epochs for distilled and 5 for synthetic; loss on the assistant turn only). Epoch budgets followed established best practice for each strategy rather than being matched across strategies; the rationale is given in the Supplement (Supplementary Methods S1). Primary results are reported at the largest matched training size (n = 2,000), so comparisons hold the training-data budget constant across configurations; results with each configuration's training size chosen by validation macro-F1 are reported as a robustness analysis (Table S1).

## Baseline models

Two zero-shot baselines bracketed the performance range on the same 100 test reports: (1) GPT-4o zero-shot, the same hosted model that produced the distilled training labels, queried without chain-of-thought against the four-class schema (the hosted-model anchor); and (2) Gemma-3-12B-it zero-shot, the instruction-tuned base model before any task-specific fine-tuning (the open-weight floor). The Gemma baseline emits finer-grained columns (separate acuity-name and density indicators), which we collapsed to the four-class schema by inclusive disjunction (e.g., acute/hyperacute/hyperdense = acute $\vee$ hyperacute $\vee$ hyperdense), matching the schema all fine-tuned models were trained against.

## Compute environment

All fine-tuning ran on a single NVIDIA L40S GPU (48 GB). Because the recipe combines 4-bit NF4 quantisation, LoRA adapters, and gradient checkpointing (via Unsloth), the loaded model, adapters, and CUDA context reserved approximately 12 GB before training (measured by torch.cuda.max_memory_reserved()); estimated peak memory during training was roughly 12–14 GB for CH and 15–20 GB for IFT, so both fit within a single 24 GB consumer GPU such as an NVIDIA RTX 4090 or RTX 3090; these peaks are analytical estimates from the pre-training footprint plus expected activation, gradient, and optimizer-state contributions, as a measured peak was not logged. Wall-clock training times are reported in Results (Training time and cost).

### Evaluation metrics

Performance was reported at the report level. The primary outcome was macro-averaged F1 across the four acuity labels. Secondary outcomes included micro-F1, weighted F1, exact-match (subset) accuracy, Hamming accuracy, per-label F1, precision, recall, and, for the CH configurations (whose sigmoid outputs are probabilistic), per-label AUROC; for IFT and both zero-shot baselines, which emit categorical tokens, AUROC could not be computed. Calibration of both CH models was assessed by per-label Brier score and reliability diagrams in five quantile bins, and per-label decision thresholds were additionally tuned by F1-maximisation on the validation set and re-evaluated on the test set. Specificity in the absence of true findings was quantified by counting the ICH-negative test reports (n = 19) receiving at least one false-positive acuity label, a direct measure of the clinically undesirable "hallucination" mode.

### Statistical analysis

95% confidence intervals were computed by paired non-parametric bootstrap (1,000 resamples of report indices, preserving within-report label correlations; 2.5th and 97.5th percentiles reported). Between-configuration comparisons used McNemar's exact test on report-level multi-label exact-match correctness (all four labels correct). Differences in macro-F1 were reported as paired bootstrap estimates with 95% CIs from resampling the same report indices for both configurations. Because these two analyses target different quantities: the bootstrap CI is computed on continuous macro-F1, whereas McNemar's test operates on the binary report-level exact-match outcome. They can diverge, and we note where they do. Sample-size sensitivity was characterised by an ordinary-least-squares fit of macro-F1 on $\log_{10}(N)$ per configuration, with slope CIs from a parametric bootstrap over the five fitted points. Analyses used Python 3.10 (scikit-learn, SciPy, pandas, NumPy).

### Cost analysis

We priced the pipeline end-to-end to compare hosted-API inference against on-premises deployment. Token counts were measured with the GPT-4o tokenizer (tiktoken, o200k_base): label-inference tokens on all 100 test reports using the deployed prompt and JSON output format, and synthetic-generation tokens on the generation prompt (system prompt, three exemplars, 30-label guide, and schema). API cost per call was (input tokens / 1M) × $2.50 + (output tokens / 1M) × $10.00 at 2026 GPT-4o standard rates, with a 50% reduction for the Batch API. Local inference cost per report was the sum of amortised hardware cost (list price divided by 3 years × 4,000 h/year) and power cost (thermal design power × $0.12/kWh at the U.S. commercial average), divided by throughput, estimated at 4,500 reports/hour on the NVIDIA L40S from published Unsloth benchmarks for Gemma-3-12B in 4-bit NF4 with LoRA. One-time build cost comprised labeling the 2,000-report distilled pool and a single fine-tuning run for the deployed model; break-even annual volume was the one-time cost, amortised over a three-year horizon, divided by the per-report API-minus-local difference. Full assumptions are in Table 4.

***Table 1***

| Label | Positive | Negative |
|---|---|---|

| ICH-present (overall) | 81 | 19 |
|---|---|---|
| Acute / hyperacute / hyperdense | 40 | 60 |
| Subacute / isodense | 5 | 95 |
| Chronic / hypodense | 6 | 94 |
| Acute-on-chronic / mixed | 11 | 89 |

*Table 1. Test-set label prevalence on the 100 manually adjudicated head-CT reports. Counts are reports containing at least one mention of each acuity class. ICH-present is the overall report-level flag; acuity labels apply to ICH-positive reports.*

## Results

### Aggregate performance and primary outcome

At n = 2,000, DIFT achieved a test macro-F1 of 0.845 [0.777, 0.908], statistically indistinguishable from GPT-4o zero-shot (0.850 [0.778, 0.911]; McNemar $p = 1.000$; Δ macro-F1 −0.005 [−0.021, +0.011]) and significantly above the Gemma-3-12B-it base model (0.667 [0.555, 0.747]; Δ +0.178 [+0.103, +0.263]; $p < 0.001$). DCH reached 0.792 [0.673, 0.862], significantly below GPT-4o (Δ −0.058 [−0.148, −0.009]; $p = 0.013$) and, on the report-level exact-match test, not significantly above the base model (Δ macro-F1 +0.125 [+0.052, +0.204]; McNemar $p = 0.108$). Neither synthetic model exceeded the base-model floor at any training size: at n = 2,000, SIFT reached 0.645 [0.523, 0.729] (Δ vs base −0.022 [−0.141, +0.085]; $p = 0.229$) and SCH reached 0.635 [0.480, 0.739] (Δ −0.032 [−0.129, +0.049]; $p = 0.065$). Learning curves across sizes are shown in Figure 2; paired comparisons are in Table 2.

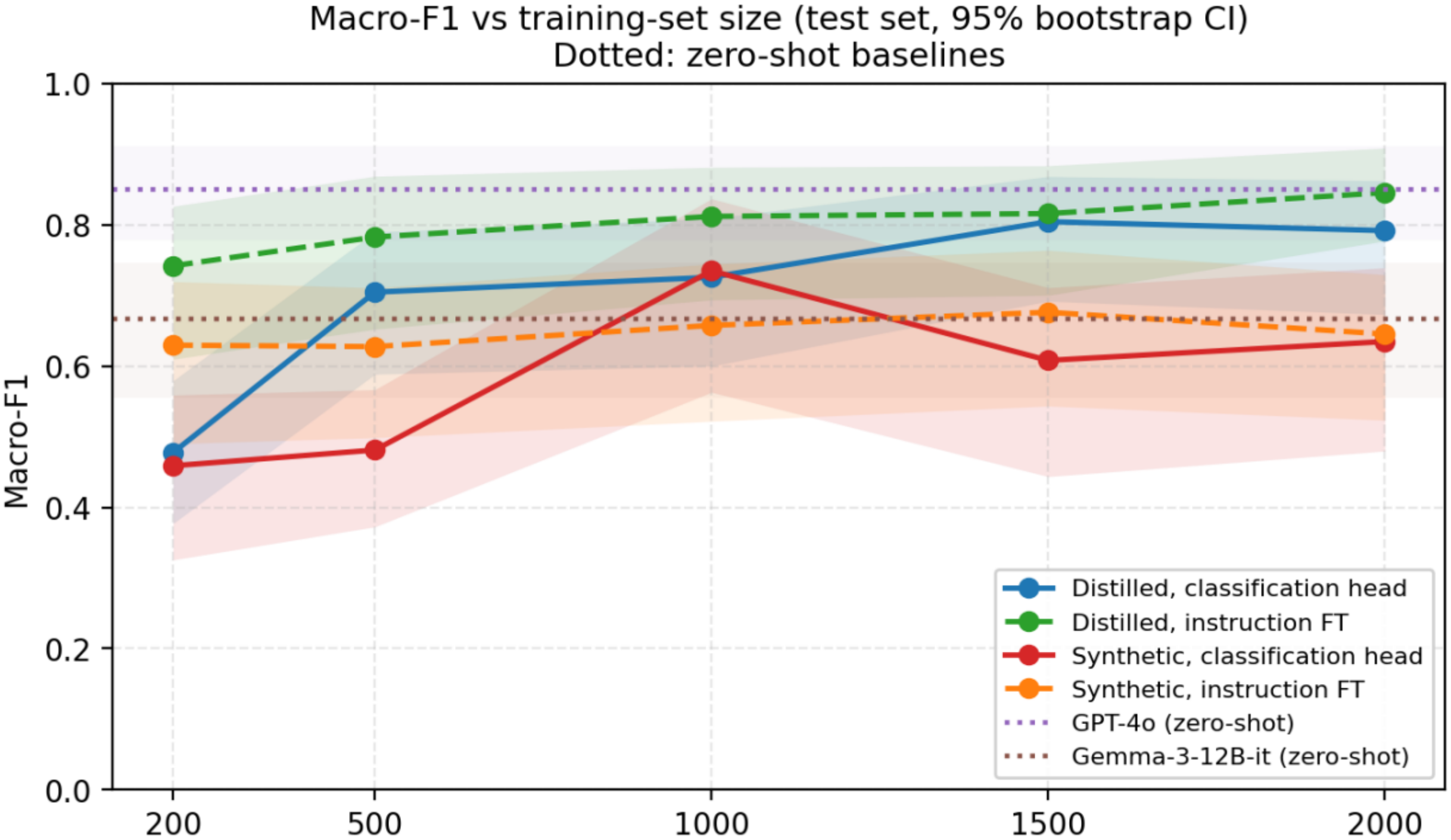

*Figure 2. Test-set macro-F1 versus training-set size for the four fine-tuned configurations (solid = classification head; dashed = instruction fine-tuning), with 95% bootstrap CI ribbons and both zero-shot baselines as horizontal reference lines. DIFT converges onto GPT-4o at n = 2,000; both synthetic curves remain at or below the open-weight floor.*

***Table 2a***

| Configuration | n | Macro-F1 (95% CI) | Exact | ICH-neg FP |
|---|---|---|---|---|
| Gemma-3-12B-it (zero-shot, base) | — | 0.667 [0.555, 0.747] | 0.66 | 1 / 19 |
| **GPT-4o (zero-shot, teacher)** | **—** | **0.850 [0.778, 0.911]** | **0.86** | **0 / 19** |
| **DIFT (distilled, instruction FT)** | **2,000** | **0.845 [0.777, 0.908]** | **0.85** | **0 / 19** |
| DCH (distilled, classification head) | 2,000 | 0.792 [0.673, 0.862] | 0.75 | 4 / 19 |
| SIFT (synthetic, instruction FT) | 2,000 | 0.645 [0.523, 0.729] | 0.58 | 2 / 19 |
| SCH (synthetic, classification head) | 2,000 | 0.635 [0.480, 0.739] | 0.54 | 8 / 19 |

*Table 2a. Headline test-set performance at n = 2,000 for all fine-tunes. Macro-F1 95% CIs from 1,000 paired bootstrap resamples. ICH-neg FP = number of the 19 ICH-negative reports receiving at least one false-positive acuity label. Bold: the statistical tie between GPT-4o and the best fine-tune (DIFT).*

***Table 2b***

| Comparison (A vs B) | Δ macro-F1 (A − B) [95% CI] | McNemar p |
|---|---|---|
| **Data-source effect (adaptation strategy held fixed)** | | |
| DCH vs SCH | +0.157 [+0.065, +0.277] | <0.001 |
| DIFT vs SIFT | +0.200 [+0.110, +0.315] | <0.001 |
| **Adaptation-strategy effect (data source held fixed)** | | |
| DCH vs DIFT | −0.054 [−0.141, −0.006] | 0.031 |
| SCH vs SIFT | −0.011 [−0.145, +0.116] | 0.585 |
| **Fine-tuned models vs GPT-4o (zero-shot)** | | |
| DIFT vs GPT-4o | −0.005 [−0.021, +0.011] | 1.000 |
| DCH vs GPT-4o | −0.058 [−0.148, −0.009] | 0.013 |
| SIFT vs GPT-4o | −0.205 [−0.321, −0.114] | <0.001 |
| SCH vs GPT-4o | −0.215 [−0.350, −0.102] | <0.001 |
| **Fine-tuned models vs Gemma-3-12B-it (zero-shot)** | | |
| DIFT vs Gemma | +0.178 [+0.103, +0.263] | <0.001 |
| DCH vs Gemma | +0.125 [+0.052, +0.204] | 0.108 |
| SIFT vs Gemma | −0.022 [−0.141, +0.085] | 0.229 |
| SCH vs Gemma | −0.032 [−0.129, +0.049] | 0.065 |
| **Baseline comparison** | | |
| GPT-4o vs Gemma | +0.183 [+0.109, +0.266] | <0.001 |

*Table 2b. Paired comparisons, grouped by effect. Δ macro-F1 is A − B (positive favors A) with 95% bootstrap CI; McNemar exact test on report-level exact-match. The two comparisons in which both factors differ (DCH vs SIFT, DIFT vs SCH) were not tested, as a difference could not be attributed to a single factor. Because macro-F1 CIs and McNemar tests target different quantities, they may diverge (e.g., DCH vs Gemma). All fine-tunes at n = 2,000.*

## Per-label performance

Per-label F1 by training-set size is shown in Figure 3 and summarised in Table S2. The dominant pattern is by data source, not by class: the distilled models (DIFT, DCH) tracked GPT-4o across every acuity class, whereas the synthetic models (SIFT, SCH) underperformed the distilled models on all four classes. On acute/hyperacute, F1 was DIFT 0.920, DCH 0.857, GPT-4o 0.920, and the un-tuned base 0.864, versus SIFT 0.800 and SCH 0.705; across all configurations, including the teacher, recall was 1.000 while precision was the bottleneck (0.78–0.85), indicating systematic over-calling rather than missed bleeds, consistent with training exclusively on ICH-positive reports (see Discussion). On AOC, DIFT, DCH, and GPT-4o each reached F1 1.000, versus SIFT 0.957 and SCH 0.900. On the low-prevalence classes, subacute F1 was DIFT 1.000, DCH 0.909, GPT-4o 1.000, and base 0.471, versus SIFT 0.500 and SCH 0.533; chronic F1 was DIFT 0.462, DCH 0.400, GPT-4o 0.480, and base 0.333, versus SIFT 0.324 and SCH 0.400,

chronic being the weakest class for every configuration including the teacher. Subacute and chronic rest on only 5 and 6 positive test cases, so their per-class intervals are wide. Relative to the un-tuned base, the synthetic models' lower aggregate macro-F1 was driven by the higher-prevalence classes (acute, AOC), where they fell below the base, rather than by the low-prevalence classes (subacute, chronic), where they matched or slightly exceeded it. The training size at which each configuration's per-label F1 peaked was DIFT at n = 2,000, DCH at n = 1,500, SIFT at n = 1,500, and SCH at n = 1,000.

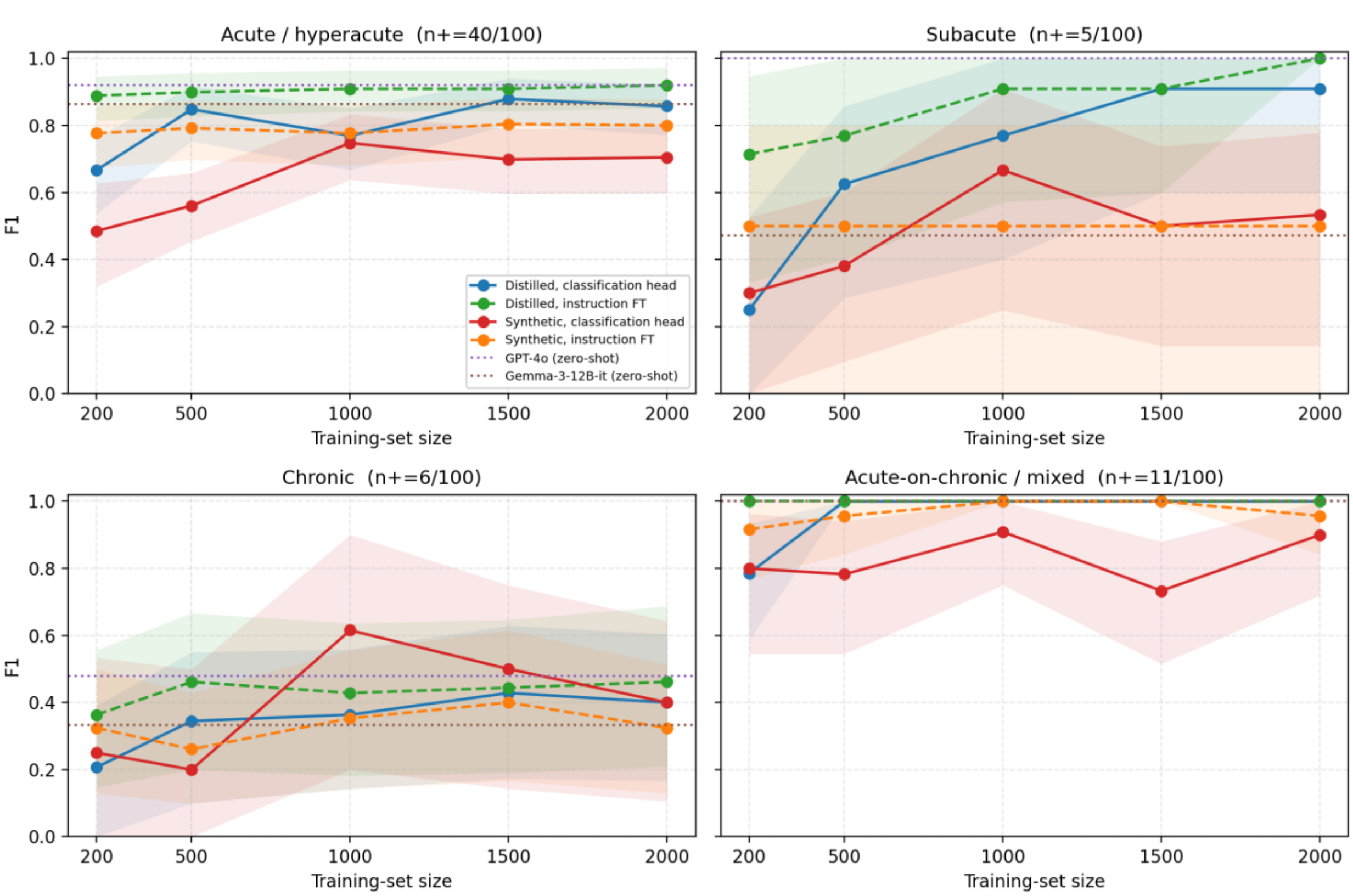

*Figure 3. Per-label F1 versus training-set size (four panels, one per acuity class; solid = CH, dashed = IFT; dotted = zero-shot baselines; shaded = 95% bootstrap CI). Intervals are wide for subacute and chronic, which have only 5 and 6 test positives.*

### Specificity on ICH-negative reports

Of the 19 ICH-negative test reports (on which all acuity labels should be zero), GPT-4o zero-shot and DIFT each produced zero false-positive labels. The Gemma-3-12B-it base model produced one, DCH four, and SIFT two. SCH was worst, with false positives on 8 of 19 reports (10 labels: 6 acute/hyperacute, 3 chronic, 1 mixed). The ordering again tracks the data source rather than the adaptation strategy.

### Discrimination and calibration of the classification-head models

Both CH models admit ROC analysis. DCH achieved macro-AUROC 0.968 [0.948, 0.985], with per-label AUROC of 0.931 (acute), 0.994 (subacute), 0.947 (chronic), and 1.000 (AOC); SCH reached 0.923 [0.881, 0.959] (Figure 4). Brier scores were uniformly better for DCH than SCH (e.g., acute/hyperacute 0.118 vs 0.252), and reliability diagrams showed positive-direction over-confidence for SCH (Supplementary Figure S1). F1-maximising per-label thresholds departed from the default 0.5 (DCH 0.80/0.05/0.85/0.05; SCH 0.95/0.65/0.95/0.05); tuning improved SCH (macro-F1 0.635→0.671; exact match 0.540→0.660) but had little effect on DCH (0.792→0.771; exact match 0.750→0.760).

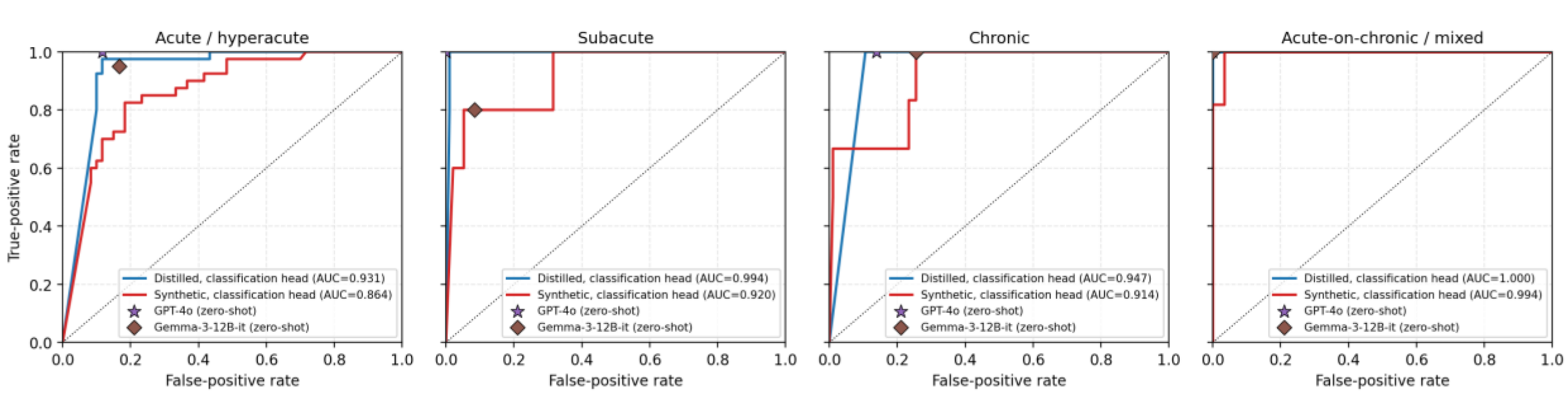

*Figure 4. Per-label ROC curves on the test set at n = 2,000 for the two CH configurations (blue = DCH, red = SCH). Operating-point markers show the categorical baselines (star = GPT-4o, diamond = Gemma-3-12B-it).*

### Sample-size sensitivity

A log-linear fit of test macro-F1 on $\log_{10}(N)$ gave slopes (95% bootstrap CI) of +0.310 [+0.134, +0.402] for DCH, +0.096 [+0.069, +0.112] for DIFT, +0.213 [−0.405, +0.434] for SCH (wide CI, non-monotonic), and +0.035 [−0.041, +0.102] for SIFT. DIFT was near asymptote at the largest size, whereas DCH was still climbing and might benefit from more distilled data; the synthetic configurations did not benefit reliably from additional data within the range tested.

### Validation-selected model performance

With training sizes chosen by validation macro-F1 (Table S1), test performance was 0.812 [0.694, 0.881] for DIFT (n = 1,000), 0.792 [0.673, 0.862] for DCH (n = 2,000), 0.657 [0.522, 0.744] for SIFT (n = 1,000), and 0.735 [0.563, 0.837] for SCH (n = 1,000). The qualitative ranking was unchanged: GPT-4o ≈ DIFT > DCH > Gemma-3-12B-it base ≈ synthetic configurations.

### Training time and cost

Wall-clock training times per run are in Table 3. IFT was far faster than CH because it used a fixed 3–5 epoch schedule versus up to 40 epochs with early stopping for CH. DIFT was the shortest to train (10m59s at n = 200 to 1h39m at n = 2,000) and SCH the longest (up to 21h53m at n = 2,000); aggregate training across all 20 experimental runs was 95 GPU-hours on a single L40S.

On a per-report basis, local inference was far cheaper than the hosted API: approximately $0.00016 per report on the L40S versus $0.0024 (standard) or $0.0012 (Batch API) for GPT-4o, roughly 8–15 times lower (Table 4). Standing up the local pipeline is a one-time cost of about $6 on the L40S (labeling 2,000 reports plus the deployed DIFT run at n = 2,000; $4.88 + $1.17), giving a break-even of roughly 900 reports/year against the standard API and 1,900/year against the Batch API, thresholds an active head-CT service readily exceeds. Charging the entire 20-run experimental sweep (95 GPU-hours, $67) to deployment raises these break-evens to roughly 10,000 and 23,000 reports/year, respectively. The

synthetic recipe is also more expensive to build than distillation ($33 vs $5 in one-time GPT-4o generation and labeling) while performing worse.

***Table 3***

| n | DCH | DIFT | SCH | SIFT |
|---|---|---|---|---|
| 200 | 1h 32m | 10m 59s | 1h 59m | 12m 37s |
| 500 | 3h 18m | 28m 01s | 4h 48m | 29m 26s |
| 1000 | 8h 45m | 49m 58s | 8h 09m | 58m 57s |
| 1500 | 10h 05m | 1h 15m | 11h 50m | 1h 26m |
| 2000 | 13h 54m | 1h 39m | 21h 53m | 1h 57m |

*Table 3. Training wall-clock per run (Weights & Biases logs), reflecting each strategy's recipe (CH: up to 40 epochs with early stopping; IFT: fixed 3 epochs distilled, 5 synthetic). Single NVIDIA L40S; per-epoch throughput was comparable across strategies (~30–35 min/epoch at n = 2,000).*

***Table 4***

| Line item | Value | Assumption / source |
|---|---|---|
| **Token measurements (GPT-4o tokenizer, o200k_base)** | | |
| Label-inference input (mean/call) | 808 tokens | 100 held-out test reports |
| Label-inference output (mean/call) | 42 tokens | fixed 4-label JSON |
| Synthetic-generation input (mean/call) | 3,373 tokens | prompt + 3 exemplars + 30-label guide + schema |
| Synthetic-generation output (mean/call) | 788 tokens | generated report + label JSON |
| **Per-report cost** | | |
| GPT-4o label inference — standard | $0.0024 | $2.50/M in + $10.00/M out |
| GPT-4o label inference — Batch API | $0.0012 | 50% discount |
| Local inference — L40S | $0.00016 | $8,000; 4,000 h/yr, 3-yr amort; 350 W @ $0.12/kWh; 4,500 rep/hr |
| GPT-4o synthetic generation — standard | $0.0163 | per generated report |
| GPT-4o synthetic generation — Batch API | $0.0082 | 50% discount |
| **One-time build cost (distilled recipe, L40S)** | | |
| Label 2,000-report distilled pool | $4.88 std / $2.44 batch | per-report label cost × 2,000 |
| Fine-tune deployed model (1 DIFT run, n = 2,000) | $1.17 | 1.65 GPU-h × amortised $/GPU-h |

| Total one-time to deploy | ≈ $6.05 | labeling + one fine-tune |
|---|---|---|
| Memo: full experimental sweep (20 runs, 95 GPU-h) | $67.32 | study compute, not deployment |
| Memo: synthetic pool generation (2,000), if used | $32.62 std / $16.31 batch | replaces distillation |
| **Break-even annual volume (3-year horizon)** | | |
| Deployed recipe vs GPT-4o standard | ≈ 900 reports/yr | one-time ≈ $6, amortised |
| Deployed recipe vs Batch API | ≈ 1,900 reports/yr | one-time ≈ $6, amortised |
| Conservative (full 95 GPU-h sweep) vs standard / batch | ≈ 10,500 / 22,600 reports/yr | one-time ≈ $72, amortised |

*Table 4. End-to-end cost model for L40S deployment (RTX-class consumer GPUs give similar or lower local cost). API priced at 2026 GPT-4o rates ($2.50/M input, $10.00/M output; Batch API 50% lower). Local cost amortises hardware over 3 years at 4,000 h/year plus power at $0.12/kWh, divided by throughput (4,500 reports/hour from Unsloth Gemma-3-12B 4-bit NF4 LoRA benchmarks; conservative given the measured 808-token label input). One-time build cost reflects the deployed distilled recipe (label 2,000 reports + one fine-tuning run); the full 20-run experimental sweep is shown as a memo. Break-even is one-time cost amortised over 3 years divided by the per-report API-minus-local difference.*

## Discussion

A fine-tuned open-weight model matched a leading hosted model at ICH-acuity extraction, but the training-data source mattered more than the fine-tuning method. At n = 2,000, DIFT was statistically indistinguishable from GPT-4o and improved on the open-weight base by 0.178 macro-F1, whereas both synthetic models failed to exceed the base-model floor at any size, despite generation being seeded with real reports as in-context exemplars. Among distilled configurations, IFT modestly outperformed CH (Δ +0.054; p = 0.031). This pattern is consistent with the view that distilling a single, well-specified behavior is far more tractable than broadly imitating a model, the latter a "false promise" [22]. Relative to the un-tuned base, distillation improved per-class F1 across all four acuity classes, whereas synthetic fine-tuning did not and on the higher-prevalence classes slightly reduced it; we did not find the synthetic deficit to be localised to the rare acuity classes. One possible contributor is coverage-narrowing, the tendency of LLM-generated text to under-represent the tails of the source distribution [28,29]; however, with only 5 and 6 positive test cases for subacute and chronic, our data can neither confirm nor exclude this, and we raise it only as a hypothesis for adequately powered future work.

### Generative versus discriminative fine-tuning

With distilled data, IFT edged out CH on macro-F1 and produced zero false positives on ICH-negative reports (matching GPT-4o), while CH produced four. CH nonetheless yields probabilistic outputs supporting discrimination and calibration analysis (DCH macro-AUROC 0.968). Strong ranking (AUROC) coexisted with weaker operating-point performance (F1) because the default 0.5 threshold was poorly calibrated; modern classifiers are commonly miscalibrated [30], and per-label threshold tuning helped SCH but not the already-strong DCH. IFT suits a categorical, deployment-ready labeler with high specificity; CH suits applications needing probabilities or tunable operating points. DCH had not plateaued (slope +0.310), whereas DIFT was near asymptote, so DCH may narrow the gap with additional distilled data.

### Per-label behavior and clinical safety

Across configurations, including the teacher, acute hemorrhage showed recall 1.000 with precision 0.78–0.85, i.e., over-calling rather than missed bleeds. This profile is favorable for screening or cohort retrieval but implies positive labels need confirmation before high-stakes use; it is also expected given that training used only ICH-positive reports, which provides no negative acuity contrast and plausibly inflates recall at the expense of precision. Chronic/hypodense was hardest for every model, including GPT-4o (F1 0.480). The dominant failure mode was conflation of chronic hemorrhage with other hypodense findings (old infarcts, encephalomalacia, post-surgical change); because the teacher, our performance ceiling, also fails here, part of this reflects intrinsic ambiguity in the reports rather than only a fixable prompting issue, although targeted prompt engineering or additional annotation of chronic cases may still help. Reassuringly, false-positive labeling on ICH-negative reports was eliminated by DIFT (0 of 19, equal to GPT-4o).

### Clinical and operational implications

For this specific task, a small open-weight model fine-tuned with modest resources reached parity with a leading hosted model. All configurations fit within a 24 GB consumer GPU (~12 GB reserved pre-training; ~12–20 GB peak), and inference runs locally in 4-bit quantisation. On-premises deployment offers three advantages: protected health information never leaves the institution; there are no per-query API fees; and the model is frozen and version-controlled, avoiding silent drift and vendor deprecation [17,18]. These properties suit high-volume, sustained uses such as continuous post-deployment surveillance and registry population, where subgroup monitoring requires a steady, private supply of reference labels [25-27]. Although the synthetic approach underperformed here, it retains a distinct privacy advantage (the generated corpus is shareable and non-identifiable) that may justify further work despite the current accuracy cost. A formal cost comparison of API versus amortized local inference is reported in Results (Training time and cost).

### Limitations

This study has limitations. Evaluation used 100 test reports from a single institution; two classes had very low positive support (subacute n = 5, chronic n = 6), giving wide intervals, and external multi-site validation is needed. Inter-annotator agreement was not quantified. We evaluated one base model (Gemma-3-12B) and one teacher (GPT-4o); because the student is distilled from GPT-4o, the teacher is also the ceiling and its errors (e.g., on chronic) are inherited. Training used only ICH-positive reports, matching the intended hierarchical use but limiting conclusions about negative discrimination. Our synthetic design is one of many (three-exemplar prompting, per-class sampling, temperature 0.4–0.8); diversity-aware or hybrid schemes might perform better, so this is not a blanket verdict against synthetic data. AUROC was available only for CH models, and we did not test chain-of-thought or extended-context prompting.

### Future directions

The most important next step is prospective, multi-institutional validation, ideally embedded in a live monitoring pipeline that performs the subgroup analysis this work is meant to enable. Because DCH had not plateaued, scaling distilled data is a low-risk way to probe the remaining gap, as is targeted enrichment of low-prevalence classes. On the data-generation side, diversity-aware synthetic prompting or blending a distilled core with synthetic augmentation may recover some of the coverage lost by synthetic generation. Extending the approach to additional findings and modalities would test whether the parity shown here for ICH acuity generalizes to other narrow report-labeling tasks.

### Conclusion

For ICH-acuity extraction from head-CT reports, an open-weight model fine-tuned on distilled real reports matched GPT-4o on local hardware, whereas exemplar-seeded synthetic data did not. Whether an open model reaches parity depends more on the training-data source than on the fine-tuning method, and the resulting on-premises model offers advantages in privacy, cost, and version stability that make it operationally preferable for routine and surveillance use.

## Statements and Declarations

### Funding

This study was funded, in part, by the National Institutes of Health (NIH) Agreement No. 1OT2OD032581. The views and conclusions contained in this document are those of the authors and should not be interpreted as representing the official policies, either expressed or implied, of the NIH.

This work was supported by the Emory University AI Image Extraction Core Facility, RRID:SCR_026693.

Dr. Gichoya is a 2022 Robert Wood Johnson Foundation Harold Amos Medical Faculty Development Program awardee and declares support from the Lacuna Fund (#67), the Gordon and Betty Moore Foundation, the NIH (NIBIB) MIDRC grant under contracts 75N92020C00008 and 75N92020C00021, NHLBI Award Numbers R01HL167811 and R01HL177003, and NIH common fund awards 1OT20D038065-01 and 1R25OD039834-01.

### Competing Interests

The authors declare no competing interests.

### Author Contributions

*[Draft — please verify each author's CRediT roles before submission; assignments below are a starting template.]*

A.M.: Conceptualisation, study design, methodology, data collection, software, validation, formal analysis, writing (original draft), review & editing; accessed and verified the data. K.M.: Software, methodology, formal analysis, writing (review & editing). M.C.: Conceptualisation, methodology, data collection, data curation, formal analysis, visualisation, writing (review & editing). J.M.: Data collection, data curation, writing (review & editing). T.D.: Data collection, validation, writing (review & editing). F.L.:

Data collection, validation, writing (review & editing). R.I.: Data collection, data curation, writing (review & editing). B.B-M.: Investigation, data collection, writing (review & editing). C.R.S.: Data collection, writing (review & editing). Y.J.: Data curation, validation, writing (review & editing). J.W.G.: Supervision, methodology, study design, data interpretation, resources, writing (review & editing). A.E.: Supervision, methodology, software, data interpretation, writing (review & editing). H.T.: Supervision, conceptualisation, study design, funding acquisition, project administration, data interpretation, writing (review & editing); accessed and verified the data. All authors reviewed the manuscript.

### Ethics approval and consent to participate

Not applicable; this work uses publicly released datasets only.

### Data Availability

The datasets generated and/or analyzed during the current study are not publicly available due to patient privacy protections (protected health information) and institutional data-sharing policy, but are available from the corresponding author on reasonable request.

### Code Availability

The code generated and/or analyzed during the current study is not publicly available due to institutional policy but is available from the corresponding author on reasonable request.

## Acknowledgements

The authors thank Wasif Bala, Hanzhou Li, John T. Moon, Chad Robichaux, Dan I. G. Cohen-Addad, and Ninad V. Salastekar for their contributions to this work.

## References

1. Irvin J, Rajpurkar P, Ko M, et al. CheXpert: A Large Chest Radiograph Dataset with Uncertainty Labels and Expert Comparison. Proc AAAI Conf Artif Intell. 2019;33:590–597. arXiv:1901.07031.

2. Labeling Noncontrast Head CT Reports for Common Findings Using Natural Language Processing. AJNR Am J Neuroradiol. 2022. doi:10.3174/ajnr.A7500.

3. Text Analysis of Radiology Reports with Signs of Intracranial Hemorrhage on Brain CT Scans Using the Decision Tree Algorithm. 2023. PMC10171057.

4. Smit A, Jain S, Rajpurkar P, Pareek A, Ng AY, Lungren MP. CheXbert: Combining Automatic Labelers and Expert Annotations for Accurate Radiology Report Labeling Using BERT. Proc EMNLP. 2020. arXiv:2004.09167.

5. Automated labelling of radiology reports using natural language processing: Comparison of traditional and newer methods. 2024. PMC11080679.

6. Automated Radiology Report Labeling in Chest X-Ray Pathologies: Development and Evaluation of a Large Language Model Framework. 2025. PMC11970564.

7. Using BERT Models to Label Radiology Reports (editorial). Radiol Artif Intell. 2022. doi:10.1148/ryai.220124.

8. Automatic text classification of actionable radiology reports of tinnitus patients using BERT and in-domain pre-training. 2022. PMC9338483.

9. In-Context Learning with Large Language Models: A Simple and Effective Approach to Improve Radiology Report Labeling. Healthc Inform Res. 2025.

10. Two-stage large language model approach enhancing entity classification and relationship mapping in radiology reports. Sci Rep. 2025. doi:10.1038/s41598-025-16213-z.

11. Vishwanath K, Alyakin A, Ghosh M, et al. General-purpose large language models outperform specialized clinical AI tools on medical benchmarks. Nat Med. 2026;32(7):2405–2409. doi:10.1038/s41591-026-04431-5.

12. Adams LC, Bressem KK, et al. Data extraction from free-text stroke CT reports using GPT-4o and Llama-3.3-70B: the impact of annotation guidelines. Acad Radiol. 2025.

13. Dorfner FJ, et al. Comparing Commercial and Open-Source Large Language Models for Labeling Chest Radiograph Reports. Radiology. 2024. doi:10.1148/radiol.241139.

14. Large Language Model-Based Uncertainty-Adjusted Label Extraction for AI Model Development in Upper Extremity Radiography. Eur Radiol. 2025. doi:10.1007/s00330-025-12102-1 (arXiv:2510.05664).

15. Synthetic data distillation enables the extraction of clinical information at scale. npj Digit Med. 2025. doi:10.1038/s41746-025-01681-4.

16. Generating synthetic clinical text with local large language models to identify misdiagnosed limb fractures in radiology reports. Artif Intell Med. 2024.

17. Chen L, Zaharia M, Zou J. How Is ChatGPT's Behavior Changing over Time? Harvard Data Sci Rev. 2024. arXiv:2307.09009.

18. Narayanan A, Kapoor S. Is GPT-4 getting worse over time? AI as Normal Technology. 2024.

19. Human-level information extraction from clinical reports with finetuned language models (Strata). 2025. PMC12749435.

20. Distilling Large Language Models for Efficient Clinical Information Extraction. 2025. arXiv:2501.00031.

21. Synthetic data trained open-source language models are feasible alternatives to proprietary models for radiology reporting. 2025. PMC12287339.

22. Gudibande A, Wallace E, Snell C, Geng X, Liu H, Abbeel P, Levine S, Song D. The False Promise of Imitating Proprietary LLMs. Proc ICLR. 2024. arXiv:2305.15717.

23. Digitalizing English-language CT Interpretation for Positive Haemorrhage Evaluation Reporting (DECIPHER). 2025. PMC12306305.

24. Chetla N, Patel S, Kumar R, Naidu T, Bouras A, Tavakkoli A, Waisberg E. Intracranial Hemorrhage Detection and Subtype Classification on CT Imaging Using a Large Language Model. Cureus. 2025;17(11):e98029. doi:10.7759/cureus.98029.

25. Statistically Valid Post-Deployment Monitoring Should Be Standard for AI-Based Digital Health. 2025. arXiv:2506.05701.

26. Distribution shift detection for the postmarket surveillance of medical AI algorithms: a retrospective simulation study. npj Digit Med. 2024. doi:10.1038/s41746-024-01085-w.

27. Diagnosing and remediating harmful data shifts for the responsible deployment of clinical AI models. 2025. PMC12138723.

28. Shumailov I, Shumaylov Z, Zhao Y, Papernot N, Anderson R, Gal Y. AI models collapse when trained on recursively generated data. Nature. 2024;631:755–759. doi:10.1038/s41586-024-07566-y.

29. How to Synthesize Text Data without Model Collapse? 2024. arXiv:2412.14689.

30. Guo C, Pleiss G, Sun Y, Weinberger KQ. On Calibration of Modern Neural Networks. Proc ICML. 2017. arXiv:1706.04599.

31. Chambon PJ, Wu C, Steinkamp JM, Adleberg J, Cook TS, Langlotz CP. Automated deidentification of radiology reports combining transformer and “hide in plain sight” rule-based methods. J Am Med Inform Assoc. 2022;30(2):318–328. doi:10.1093/jamia/ocac219.